\documentclass{article}
\usepackage{amsmath,graphicx}
\usepackage{url} 
\usepackage{booktabs}   
\usepackage{multirow}   
\usepackage{amssymb}    
\usepackage[preprint]{spconf}
\toappear{Accepted by 2026 IEEE International Symposium on Biomedical Imaging (ISBI)}
\copyrightnotice{ISBI 2026}

\usepackage{enumitem}
\setlist{nosep, leftmargin=14pt}

\title{3D Modality-Aware Pre-training for Vision-Language Model in MRI Multi-organ Abnormality Detection}

\name{Haowen Zhu$^{b}$ \qquad Ning Yin$^{c}$ \qquad Xiaogen Zhou$^{a}$\sthanks{Xiaogen Zhou is the corresponding author (xiaogenzhou@126.com).}}

\address{$^{a}$ School of Electronic, Electrical Engineering and Physics, Fujian University of Technology \\
    $^{b}$ School of Computer Science and Engineering, Southeast University, China \\
    $^{c}$ Department of Medical Imaging, Suzhou Traditional Chinese Medicine Hospital, China
}

\begin{document}
\setcounter{page}{1}
\maketitle

\begin{abstract}
Vision-language models (VLMs) show strong potential for complex diagnostic tasks in medical imaging. However, applying VLMs to multi-organ medical imaging introduces two principal challenges: (1) modality-specific vision-language alignment and (2) cross-modal feature fusion. In this work, we propose MedMAP, a \underline{Med}ical \underline{M}odality-\underline{A}ware \underline{P}retraining framework that enhances vision-language representation learning in 3D MRI. MedMAP comprises a modality-aware vision-language alignment stage and a fine-tuning stage for multi-organ abnormality detection. During the pre-training stage, the modality-aware encoders implicitly capture the joint modality distribution and improve alignment between visual and textual representations. We then fine-tune the pre-trained vision encoders (while keeping the text encoder frozen) for downstream tasks. To this end, we curated MedMoM-MRI3D, comprising 7,392 3D MRI volume–report pairs spanning twelve MRI modalities and nine abnormalities tailored for various 3D medical analysis tasks. Extensive experiments on MedMoM-MRI3D demonstrate that MedMAP significantly outperforms existing VLMs in 3D MRI–based multi-organ abnormality detection. Our code is available at \url{https://github.com/RomantiDr/MedMAP}.
\end{abstract}


\section{Introduction}
\label{sec:intro}
The analysis of 3D medical images, such as multi-modal Magnetic Resonance Imaging (MRI), is a critical yet labor-intensive task in clinical practice. While deep learning has shown promise, supervised methods are often constrained by the need for extensive, expert-level annotations for predefined disease or abnormality categories \cite{bInt_02}.\\
\indent Vision-language models (VLMs) offer a promising alternative by learning from readily available image-report pairs \cite{bInt_04}. However, existing VLMs face critical limitations when applied to 3D medical imaging diagnostic tasks. First, many successful models like MedCLIP \cite{bInt_7} and BiomedCLIP \cite{bInt_8_0} are designed for 2D images and cannot directly capture the rich spatial and anatomical context of 3D volumetric data. Second, recent 3D VLMs \cite{Exp_5, RW_20} often treat different MRI modalities (e.g., T1, T2, DWI) as modality-agnostic inputs. This overlooks the unique diagnostic information embedded in each sequence, leading to suboptimal feature representation \cite{Md_3}. Finally, most VLMs rely on coarse, global-level contrastive learning between entire volumes and reports, failing to capture fine-grained correspondences between specific anatomical regions and descriptive sentences.\\
\indent In this paper, we propose MedMAP, a fine-grained vision-language alignment framework for 3D multi-organ abnormality detection in MRI. MedMAP comprises two stages: a modality-aware pre-training (MAP) stage and a multi-organ abnormality detection fine-tuning stage. The MAP stage enables modality-level fine-grained alignment between 3D MRI volumes and radiology reports. Motivated by intrinsic correspondence in reports, where abnormal findings are documented per organ, structure, and modality, we perform modality-level decomposition and matching for both images and reports. This is followed by fine-grained alignment of matched visual and textual embeddings within the same modality, mitigating misalignment issues in global contrastive learning and improving VLM interpretability. Additionally, we introduce a cross-modal semantic aggregation (CSA) module to integrate visual and textual tokens via cross-modal interactions. Extensive experiments on the MedMoM-MRI3D dataset demonstrate the advantages of MedMAP over state-of-the-art counterparts.
\vspace{-4mm}
\section{Methodology}
\label{sec:method}
\begin{figure*}[t]
  \centering
  \centerline{\includegraphics[width=0.9\textwidth]{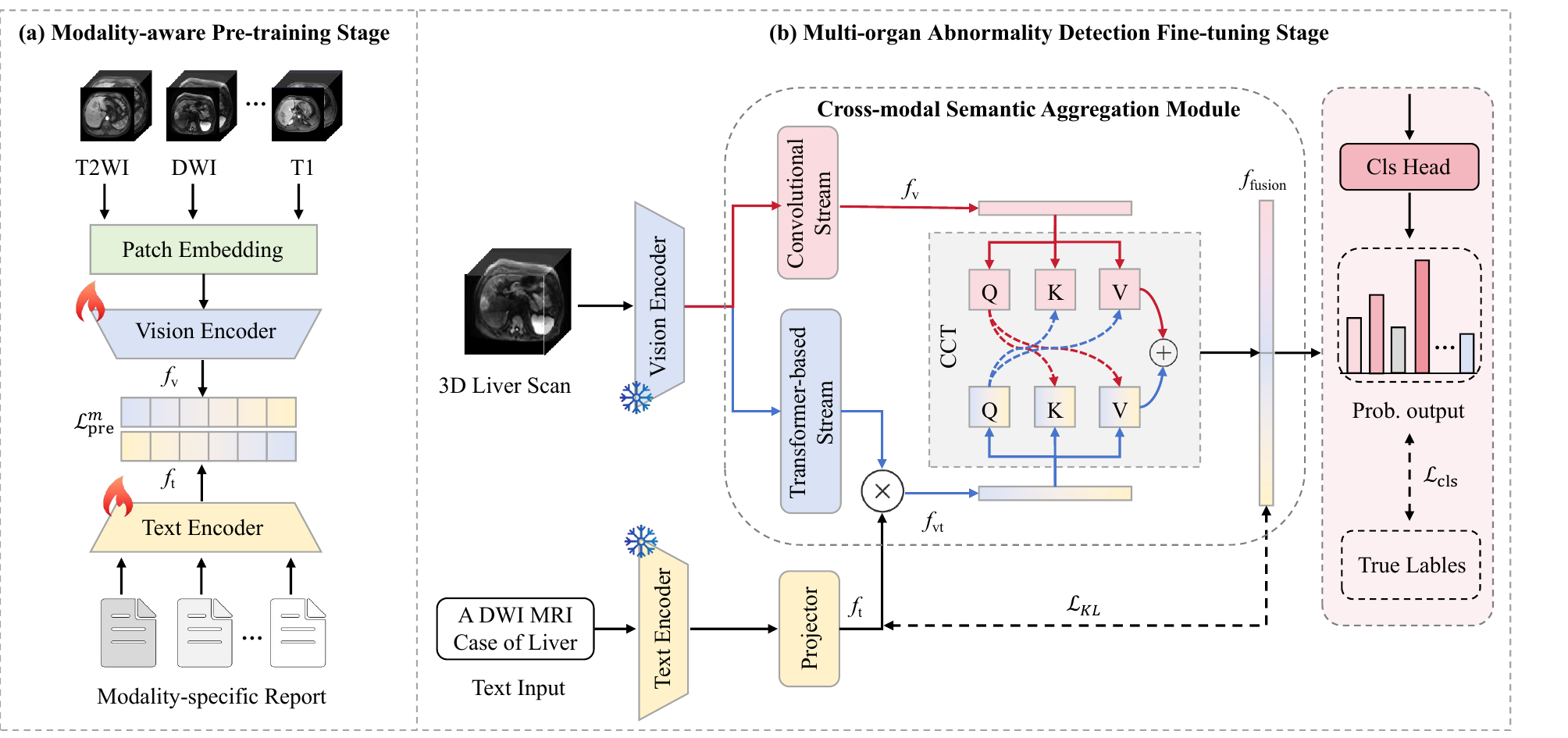}}
  \caption{Overview of the MedMAP framework.
(a) Pre-training stage: The vision and text encoders are pretrained using modality-specific MRI volumes and reports.
(b) Fine-tuning stage: The text encoder remains frozen, while a projector and the vision pipeline are trained. Additionally, the cross-modal semantic aggregation (CSA) module integrates pathological visual and textual tokens through structured cross-modal interactions.}
  \label{fig:framework}
  \vspace{-3mm} 
\end{figure*}

We propose MedMAP, a framework for 3D vision-language representation learning in multi-modal MRI. As illustrated in Fig.~\ref{fig:framework}, it consists of two main stages: a modality-aware pre-training stage and a fine-tuning stage for downstream tasks.
\subsection{Modality-aware Vision-Language Pre-training}
\label{ssec:pretrain}
To learn specialized representations, we pre-train modality-specific vision encoders. As shown in Fig.~\ref{fig:framework}(a), for a given pair $(V^m, T^m)$ of a 3D MRI volume and its corresponding report, the vision encoder $e_v$ extracts a visual feature $f_v$. Concurrently, a text encoder $e_t$ processes the report to yield a textual feature $f_t$. Both encoders are trained by optimizing a symmetric contrastive loss, $\mathcal{L}_{\text{pre}}^m$, to maximize the similarity between paired visual and textual representations. The primary goal of this stage is to equip the vision encoders with the ability to extract diagnostically relevant features that are unique to each MRI sequence. The overall pre-training loss for a specific modality $m$ is:
\begin{equation}
\label{eq:contrastive_loss}
\mathcal{L}_{\text{pre}}^m(f_v, f_t) = \frac{1}{2} \left( \mathcal{L}_{\text{v2t}}(f_v, f_t) + \mathcal{L}_{\text{t2v}}(f_t, f_v) \right)
\end{equation}
This process is repeated for each modality, creating a set of expert vision encoders.
\subsection{Fine-tuning for Multi-organ Abnormality Detection}
\label{ssec:finetune}
We propose the Cross-Modal Semantic Aggregation (CSA) module, which processes the fused representation via two parallel branches: a convolutional stream and a Transformer-based stream. This dual-path architecture is designed to capture complementary semantic features from the input data. The convolutional stream comprises a stack of 3D convolutional layers that extract robust local spatial features, resulting in a feature map denoted as $f_{v}$. In parallel, the Transformer stream consists of a series of 3D Transformer blocks based on the Swin Transformer architecture \cite{Exp_4}, which are employed to model long-range dependencies and capture global contextual information. By integrating both local and global representations, the CSA module enables more comprehensive feature modeling for multi-organ abnormality detection tasks.\\
\indent Additionally, the text encoder is kept frozen during fine-tuning, and its output is passed through a trainable projection layer to obtain a refined text embedding. This projected text feature modulates the output of the Transformer stream via element-wise multiplication, resulting in a text-guided visual representation, denoted as $f_{vt}$. The original visual feature $f_{v}$ and the text-guided feature $f_{t}$ are then fused using a Cross-Cognition Transformer (CCT) \cite{Md_3}, which employs bidirectional cross-attention to enable deep interaction between the two modalities. This mechanism facilitates a semantic–spatial interplay, where the semantic ``what" from the textual guidance interacts with the spatial ``where" from the visual stream, leading to a more robust, informative, and interpretable fused representation.

\subsubsection{Overall Objective Function}
The fine-tuning is optimized by a hybrid loss. The primary loss is the binary cross-entropy (BCE) loss $\mathcal{L}_{\text{cls}}$. To ensure vision-language semantic alignment, we add a KL-divergence loss, $\mathcal{L}_{\text{KL}}$, between the final fused feature $f_{\text{fusion}}$ and the output of the text projector. The fine-tuning losses are defined as follows:
\begin{equation} \label{cls_loss}
\small
\mathcal{\mathcal{L}}_{\text{cls}}=\mathcal{L}_{\text{BCE}}(X_{Prob.},Y_i)
\end{equation}
\begin{equation} \label{kl_loss}
\small
\mathcal{\mathcal{L}}_{\text{KL}}=\mathcal{L}_{\text{KL}}(f_t, f_{fusion}),
\end{equation}
where $X_{Prob.}$ is the category probability output from the classification head, and $Y_i$ denotes the true label for the $i_{th}$ sample. The overall objective loss function is defined as:
\begin{equation} \label{final_cls_loss}
\small
\mathcal{L}_{total}= \lambda_c \mathcal{L}_{cls} + \lambda_s  \mathcal{\mathcal{L}}_{\text{KL}},
\end{equation}
where $\lambda_c$ and $\lambda_s$ are weighting coefficients that balance the $\mathcal{\mathcal{L}}_{\text{cls}}$ and $\mathcal{\mathcal{L}}_{\text{KL}}$. These weights are adaptively modulated using a ramp-up and ramp-down scheduling strategy, defined as $\lambda_c = 0.1 \cdot e^{-5(1-\frac{t}{t_{max}})}$ and $\lambda_s = 0.1 \cdot e^{-5(\frac{t}{t_{max}})}$, where $t$ is the current training epoch and $t_{max}$ is the total number of training epochs.
\section{Experiments}
\label{sec:exp}

\subsection{Experimental Setup}
\label{ssec:setup}

\noindent\textbf{MedMoM-MRI3D Dataset.} 
We constructed a public large-scale benchmark, MedMoM-MRI3D, for 3D vision-language multi-organ medical analysis. It comprises 7,392 3D volume–report pairs, integrating and processing several public datasets (e.g., LLD-MMRI \cite{Exp_9}, RadGenome–Brain MRI \cite{Exp_11}) to cover multiple organs (e.g., liver, brain), twelve MRI modalities, and nine distinct associated abnormalities. To enrich the dataset for pre-training, we employed GPT-4o \cite{Exp_19} to generate modality-specific reports for each case. All generated texts were subsequently verified by expert radiologists.

\noindent\textbf{Implementation Details.} Our framework was implemented in PyTorch. For pre-training, each modality expert was trained for 300 epochs using the AdamW optimizer \cite{Exp_3} with a learning rate of $10^{-4}$. The shared text encoder was a frozen BioBERT \cite{Exp_1}. For fine-tuning, the full model was trained for 500 epochs with a learning rate of $10^{-5}$. All 3D MRI volumes were resized to a uniform dimension of $128 \times 128 \times 128$.

\subsection{Main Results}
\label{ssec:results}

We evaluated MedMAP against state-of-the-art (SOTA) methods on liver and brain abnormality detection tasks. As shown in Table~\ref{tab:main_results}, MedMAP achieves a SOTA on the seven-class liver abnormality detection task, with an accuracy of 91.57\% and an AUC of 88.14\%, significantly outperforming prior VLM-based approaches. It also demonstrates superior generalizability on the binary (benign vs. malignant) brain tumor detection task, reaching 90.86\% accuracy.

\begin{table}[t]
\caption{Performance comparison with SOTA methods. Best results are in bold. Acc. is Accuracy.}
\label{tab:main_results}
\centering
\small 
\setlength{\tabcolsep}{4pt} 
\begin{tabular}{l|cc|cc}
\toprule
\multirow{2}{*}{\textbf{Method}} & \multicolumn{2}{c|}{\textbf{Liver (Multi-Class)}} & \multicolumn{2}{c}{\textbf{Brain (Binary)}} \\
\cmidrule{2-5}
& Acc. (\%) & AUC (\%) & Acc. (\%) & AUC (\%) \\
\midrule
Baseline \cite{bInt_8_3}    & 82.86 & 75.15 & 81.86 & 81.33 \\
MCPL \cite{bInt_8_1}        & 87.87 & 86.49 & 88.21 & 84.53 \\
MedCLIP \cite{bInt_7}       & 85.53 & 84.21 & 85.84 & 83.64 \\
PI-RADS \cite{RW_10}      & 86.24 & 85.46 & 87.83 & 85.61 \\
\midrule
\textbf{MedMAP (Ours)} & \textbf{91.57} & \textbf{88.14} & \textbf{90.86} & \textbf{87.33} \\
\bottomrule
\end{tabular}
\vspace{-4mm} 
\end{table}

\subsection{Ablation and Qualitative Analysis}
\label{ssec:ablation}
To validate our design, we conducted ablation studies. As Table~\ref{tab:ablation_cls_result} shows, each component contributes positively to the final performance. The modality-aware vision-language pre-training (MAVLP) provides a solid foundation (+1.36\% Acc), while adding Cross-Cognition Transformer (CCT) further boosts it (+3.03\%). The largest gain comes from our CSA module (+4.32\%), highlighting the effectiveness of its dual-stream fusion architecture.

\begin{table}[h]
\caption{Ablation study of MedMAP components on the MedMoM-MRI3D-Liver dataset. ACC is Accuracy.}
\label{tab:ablation_cls_result}
\centering
\small 
\begin{tabular}{cccc|c}
\toprule
\multicolumn{4}{c|}{\textbf{Components}} & \multirow{2}{*}{\textbf{ACC (\%)}} \\ 
\cmidrule{1-4}
Baseline & MAVLP & CCT & CSA & \\ 
\midrule
$\checkmark$ & & & & 82.86 \\
$\checkmark$ & $\checkmark$ & & & 84.22 \\
$\checkmark$ & $\checkmark$ & $\checkmark$ & & 87.25 \\
$\checkmark$ & $\checkmark$ & $\checkmark$ & $\checkmark$ & \textbf{91.57} \\ 
\bottomrule
\end{tabular}
\vspace{-4mm} 
\end{table}

\begin{figure}[t]
\centering
\begin{minipage}[b]{1.0\linewidth}
  \centering
  \centerline{\includegraphics[width=0.9\linewidth]{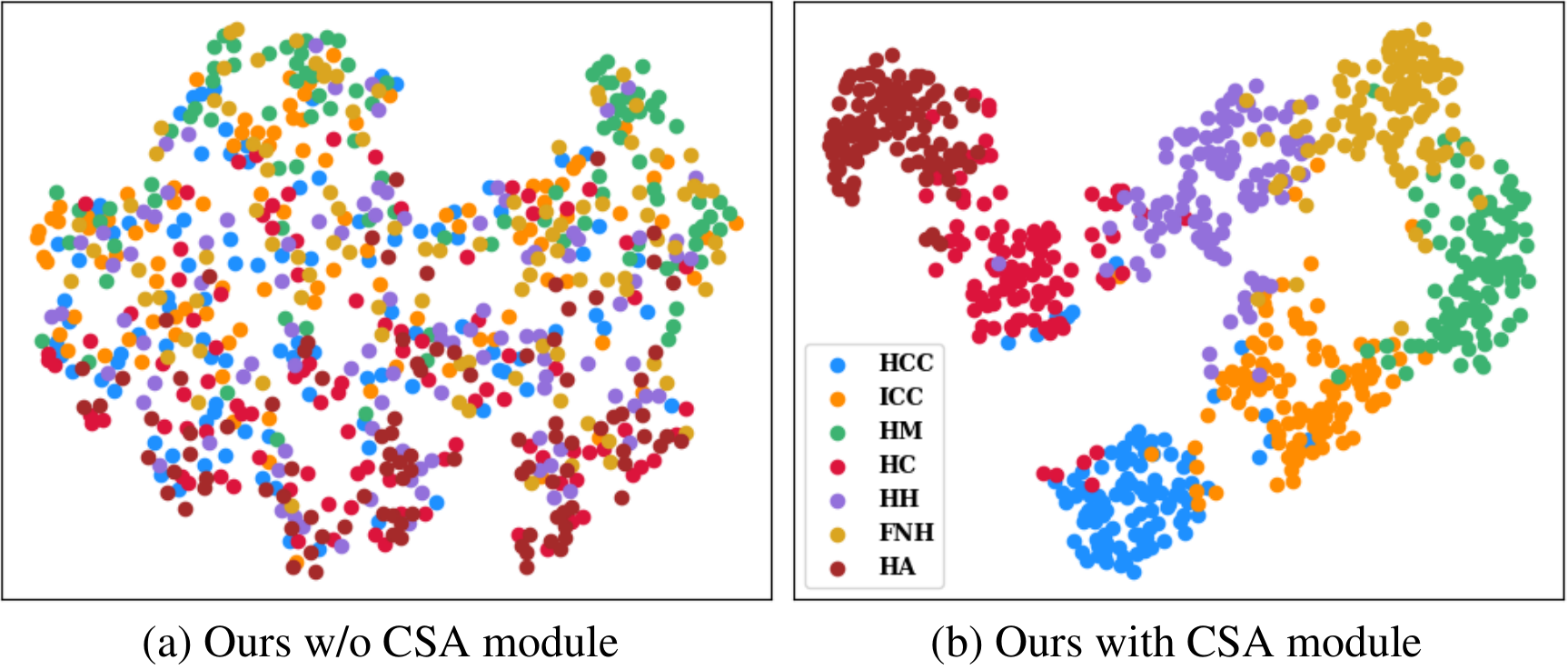}}
  \centerline{(i) t-SNE visualization of learned features.}\medskip
\end{minipage}
\begin{minipage}[b]{1.0\linewidth}
  \centering
  \centerline{\includegraphics[width=0.9\linewidth]{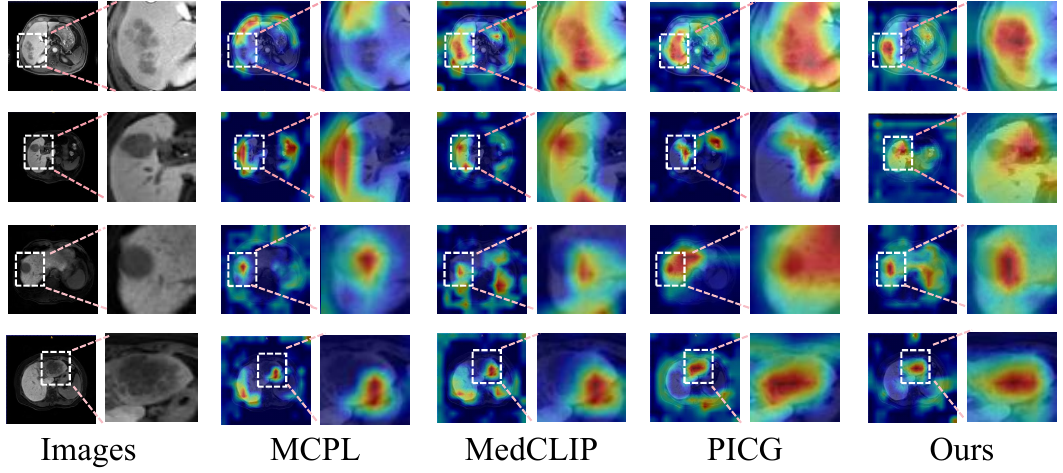}}
  \centerline{(ii) Model interpretability for liver lesion localization.}\medskip
\end{minipage}
\caption{Qualitative analysis of (i) CSA module and (ii) model interpretability.}
\label{fig:qualitative}
\vspace{-5mm} 
\end{figure}

Qualitative analysis further supports our quantitative findings. The t-SNE visualizations in Fig.~\ref{fig:qualitative}(i) show that MedMAP learns significantly more discriminative features, forming well-separated clusters compared to the baseline without the CSA module. Furthermore, the class activation maps (CAMs) in Fig.~\ref{fig:qualitative}(ii) demonstrate MedMAP's superior interpretability. Its attention is precisely focused on the pathological lesions, unlike competing methods that often produce diffuse and unfocused heatmaps. This indicates that our CSA module effectively grounds its predictions in the correct visual evidence.

\section{Conclusion}
\label{sec:conclusion}
In this paper, we present MedMAP, a vision–language framework designed for 3D multi-organ abnormality detection. By incorporating modality-aware pre-training and a novel cross-modal semantic aggregation module, MedMAP effectively learns modality-specific representations and enables deep, synergistic fusion between visual and textual modalities. Extensive experiments on the MedMoM-MRI3D benchmark demonstrate that MedMAP achieves state-of-the-art performance in both liver and brain abnormality detection, while also exhibiting improved model interpretability. In future work, we plan to extend the framework to dense prediction tasks, such as language-guided 3D medical image segmentation and reasoning, further enhancing its applicability in clinical scenarios.

\section{Compliance with ethical standards}
\label{sec:ethics}
This research study was conducted retrospectively using human subject data made available in open access by sources such as LLD-MMRI and RadGenome-Brain MRI. Ethical approval was not required as confirmed by the licenses attached with the open access data.

\section{Acknowledgments}
\label{sec:acknowledgments}
This work was supported by the National Natural Science Foundation of China (62471207).

\bibliographystyle{IEEEbib}
\bibliography{strings,refs}

\end{document}